\documentclass{article}

    \PassOptionsToPackage{numbers, compress}{natbib}

     \usepackage[preprint]{neurips_2022}

\usepackage{amsfonts}       
\usepackage{booktabs}       
\usepackage{comment}
\usepackage[T1]{fontenc}    
\usepackage[utf8]{inputenc} 
\usepackage{graphicx}    
\usepackage{hyperref}       
\usepackage{microtype}      
\usepackage{nicefrac}       
\usepackage{url}            
\usepackage{xcolor}         

\title{A Preliminary Study on Pattern Reconstruction for Optimal Storage of Wearable Sensor Data}

\author{
 Sazia Mahfuz and Farhana Zulkernine \\
School of Computing\\
Queen's University\\
Kingston, ON K7L2N8, Canada \\
\texttt{\{sazia.mahfuz, farhana.zulkernine\}@queensu.ca} \\
}

\begin{document}

\maketitle

\begin{abstract}
  Efficient querying and retrieval of healthcare data is posing a critical challenge today with numerous connected devices continuously generating petabytes of images, text, and internet of things (IoT) sensor data. One approach to efficiently store the healthcare data is to extract the relevant and representative features and store only those features instead of the continuous streaming data. However, it raises a question as to the amount of information content we can retain from the data and if we can reconstruct the pseudo-original data when needed. By facilitating relevant and representative feature extraction, storage and reconstruction of near original pattern, we aim to address some of the challenges faced by the explosion of the streaming data. We present a preliminary study, where we explored multiple autoencoders for concise feature extraction and reconstruction for human activity recognition (HAR) sensor data. Our Multi-Layer Perceptron (MLP) deep autoencoder achieved a storage reduction of 90.18\% compared to the three other implemented autoencoders namely convolutional autoencoder, Long-Short Term Memory (LSTM) autoencoder, and convolutional LSTM autoencoder which achieved storage reductions of 11.18\%, 49.99\%, and 72.35\% respectively. Encoded features from the autoencoders have smaller size and dimensions which help to reduce the storage space. For higher dimensions of the representation, storage reduction was low. But retention of relevant information was high, which was validated by classification performed on the reconstructed data.

\end{abstract}

\section{Introduction}

Streaming data is growing exponentially. Forbes reports that, by 2025, the quantity of data will double every 12 hours \citep{Forbes-March2019}. Also in one of their recent reports they stated that "RBC Capital Market projects that: by 2025, the compound annual growth rate of data for healthcare will reach 36\%" \citep{Forbes-August2021}. A high portion of the healthcare streaming data comes from the internet of things (IoT) i.e., numerous connected devices on the Internet. This explosion in IoT data has presented a critical challenge in effective data storage and management for query, analysis, and decision support \citep{Talend-June2019, Priceonmoics-June2019, Forbes-February2019}. 

Especially during the COVID peiod, tele-health, contact tracing, outbreak tracking, virus testing, remote work, and medical research have resulted in an explosion of healthcare data which has exceeded all previous estimates about growth of data \citep{Forbes-August2021}. Declining cost of storage allowed businesses and general users to store all the data. However, the exponential growth in healthcare data and the large volume of stored data is posing challenges in managing, retrieving, linking and extracting usable knowledge from the data for efficient decision support. 

The above challenges motivated us to explore effective means of knowledge representation. In this paper, we propose a method that can allow storage reduction while retaining useful information for decision support, specifically for performing a classification task. Specifically, we address three research questions, a) how can the characteristic features can be extracted from streaming data using machine learning models, b) how can the extracted representative features be stored for reducing storage, and c) how can a pseudo-original representation be reconstructed from the stored concise representative features for decision support such as performing a simple classification task?

We trained multiple autoencoder models to learn important data features while minimizing reconstruction loss. We validate our approach for Human Activity Recognition (HAR) use case scenario using streaming IoT data \citep{Anguita2013APD}.

\subsection{Contributions}
The contributions of this research work are as follows:
\begin{itemize}
    \item Storage reduction using the concise representative features instead of he whole incoming data.
    \item Validation of the usability of the stored concise data by reconstructing a pseudo-original data from the stored representative features and applying it to a classification task instead of the original data.
\end{itemize}

\subsection{Organization of the Paper}
The paper is organized into the following sections. Section \ref{Sec:Related_work} discusses the related work for storage reduction of IoT data as well as time-series data reconstruction using autoencoders. 
Section \ref{Sec:Implementation} presents the description, results, and discussion of the performed experiments. Section \ref{Sec:Threats} discusses the validity threats for this research work. Finally, section \ref{Sec:Conclusion} discusses the final thoughts and conclusion of the research work. 

\section{Related Work} \label{Sec:Related_work}

\citet{Rani-2021} discussed the different storage optimization techniques for IoT data \citep{Rani-2021}. Moreover, according to \citet{CORREA-2022}, lossy Data Compression (DC) techniques can be better alternatives to the lossless ones as they are computationally less complex as well as provide a better compression ratio. In their paper, they discussed a category of lossy compression techniques which was based on the machine learning models using artificial neural network (ANN) architectures. For our research work, we focus only on this category as there is an increasing use of ANNs in IoT scenarios to implement smart devices, paving a way to fulfilling the concept of smart cities.


\subsection{Autoencoders} \label{SubSec:Autoencoders}
Reconstructing and generating of text, image, and other types of data has come a long way since its advent due to the progress and widespread application of deep learning using ANNs. Just by itself, the area of image reconstruction in generating high-quality images from corrupted, noisy, or low-quality images has opened the door to a whole myriad of applications. Though the training of the deep learning models requires a large amount of data, now we are also living in an era of big data where data is available in abundance. We focus on autoencoders for the reconstruction of IoT data for their increasing and versatile use in reconstructing and generating other types of data like image. 

We focus on autoencoders - a well-known type of ANN for unsupervised learning - to reconstruct IoT data. Autoencoders are considered to be popular in the area of data reconstruction and data generation for different types of data such as images. We base our work on four specific types of autoencoders, namely Multi-layer perceptron (MLP) deep autoencoder, convolutional autonencoder, Long-short term memory (LSTM), and convolutional LSTM autoencoder. The justification behind using MLP deep autoencoder is its simple architecture compared to the other models. The remaining autoencoders have been chosen based on their application on time-series data.


\citet{Alaa-2019} and \citet{NGUYEN-2021} \citep{Alaa-2019,NGUYEN-2021} showed that LSTM autoencoder works better in modeling time-series data.
Again in the work of \citet{ZhangChuxu-2019}, they used multi-scale (resolution) signature matrices to characterize multiple levels of the system statuses in different time steps for their time-series data. Given the signature matrices, a convolutional encoder was employed to encode the time series correlations and an attention based Convolutional Long-Short Term Memory (ConvLSTM) network was developed to capture the temporal patterns. Finally, based upon the feature maps which encode the time-series correlations and temporal information, a convolutional decoder was used to reconstruct the input signature matrices and the residual signature matrices were further utilized to detect and diagnose anomalies. This approach was validated using synthetic dataset and real power plant dataset.

\section{Implementation} \label{Sec:Implementation}
We have saved the concise representation from the encoder part of a trained autoencoder. Four models implemented for the reconstruction component are MLP deep autoencoder, 
convolutional autoencoder, LSTM autoencoder, and convolutional LSTM autoencoder. 


The experiments have been run on Google Colab using GPU hardware accelerator in runtime. The dataset used has been UCI HAR data \citep{Anguita2013APD}. We choose this dataset  because of its simplicity and widespread use \citep{SMahfuz-2018,Niloy-2019,Hongkai-2019}. The UCI HAR dataset contains 6 activity classes (WALKING, WALKING\_UPSTAIRS, WALKING\_DOWNSTAIRS, SITTING, STANDING, LAYING). 30 volunteers have performed the six activities wearing a smartphone (Samsung Galaxy S II) on the waist. Using the smartphone's embedded accelerometer and gyroscope, 3-axial linear acceleration and 3-axial angular velocity at a constant rate of 50Hz had been captured. Pre-processing has been done by applying noise filters on the accelerometer and gyroscope signals. Then they have been sampled in fixed-width sliding windows of 2.56 sec and 50\% overlap (128 readings/window). Details of the dataset characteristics are shown in Table \ref{tab1e:UCI_Har_dataset}.
\begin{table}[t]
\caption{Details of UCI HAR dataset}
\label{tab1e:UCI_Har_dataset}
\begin{center}
\begin{tabular}{p{20mm}p{20mm}p{20mm}p{20mm}}
\textbf{No. of Subjects}  & \textbf{Age Range} &
\textbf{No. of Activities}  &\textbf{No. of Samples}
\\ 
\hline \rule{0pt}{3ex} 
30 & 19-48 & 6 & 10299\\
\bottomrule
\end{tabular}
\end{center}
\end{table}

\subsection{Validation}
The validation is done based on the classification accuracy for the reconstructed data using the convolutional LSTM classifier. We choose convolutional LSTM classifier for our validation as this model has proved to be effective in classifying time-series data \citep{ZhangChuxu-2019}.The storage reduction in percentage is considered to determine the useability of the approach for the optimum storage of the wearable sensor data.

The convolutional LSTM classifier consists of 9 layers in the following order: TimeDistributed(conv1D (64)) -> TimeDistributed (conv1D (64)) -> DropOut (0.5) -> MaxPooling1D -> Flatten -> LSTM (100) -> DropOut (0.6) -> Dense (100) -> Dense (6). 

All of the conv1D and Dense layers have ReLU as the activation function except for the last Dense layer, which has SoftMax activation function. The learning rate is set to 0.00001. Categorical cross-entropy is used as the loss function, and the batch size is set to 16. The classifier achieves a training accuracy of 96.16\% after 150 epochs.

\subsection{Experimentation}
We run the experiments as follows:
The deep autoencoders have, first, been trained on the same training data as the convolutional LSTM classifier. The concise representations have, then, been saved from the encoder layers of the deep autoencoders. The saved representations have, next, been loaded from the storage and reconstructed using the decoder parts of the autoencoders. The reconstructed representations have, finally, been fed into the convolutional LSTM classifier.

\begin{itemize}
\item \textbf{Experiment 1:}
The first experiment is performed with a simple MLP deep autoencoder. Even though its architecture is simple, it has achieved reasonable results in comparison to the other models. The MLP encoder has 5 layers consisting of 512, 256, 128, 64, 32 neurons, respectively. The MLP decoder has 5 layers consisting of 64, 128, 256, 512, 1152 neurons, respectively. All layers use Rectified Linear Unit (ReLU) as the activation function except for the last layer in the decoder which, uses Sigmoid activation function. For this experiment, the dataset has been normalized to the range between 0 and 1. The dataset has also been modified such that the sliding windows have been flattened.

The default learning rate of ADAM optimizer is 0.001. MSE is used as the loss function, and the batch size is set to 128. Fixed validation data has been used. The recorded validation loss is 0.0038 for 20 epochs.

\item \textbf{Experiment 2:}
The second experiment is performed with a convolutional deep autoencoder. The convolutional encoder consists of 6 layers in the following order: conv2D (16) -> MaxPooling2D -> conv2D (32) -> MaxPooling2D -> conv2D (64). 
\newline
The decoder consists of 7 layers in the following order: conv2D (64) -> UpSampling2D -> conv2D (32) -> UpSampling2D -> conv2D (16) -> conv2D (1). The activation function of all of the layers is ReLU except for the last layer in the decoder, which uses linear activation function. 

We use the default learning rate of ADAM optimizer, which is 0.001. MSE is used as the loss function, and the batch size is set to 16. Fixed validation data has been used. This setting has recorded a validation loss of 0.0024 after 150 epochs.

\item \textbf{Experiment 3:} 
The third experiment is performed with a LSTM deep autoencoder. Figure \ref{Fig:LSTM_autoencoder} shows the information flow between the LSTM autoencoder and the classifier.
\begin{figure}[h]
\begin{center}
\includegraphics[width=0.6\linewidth]{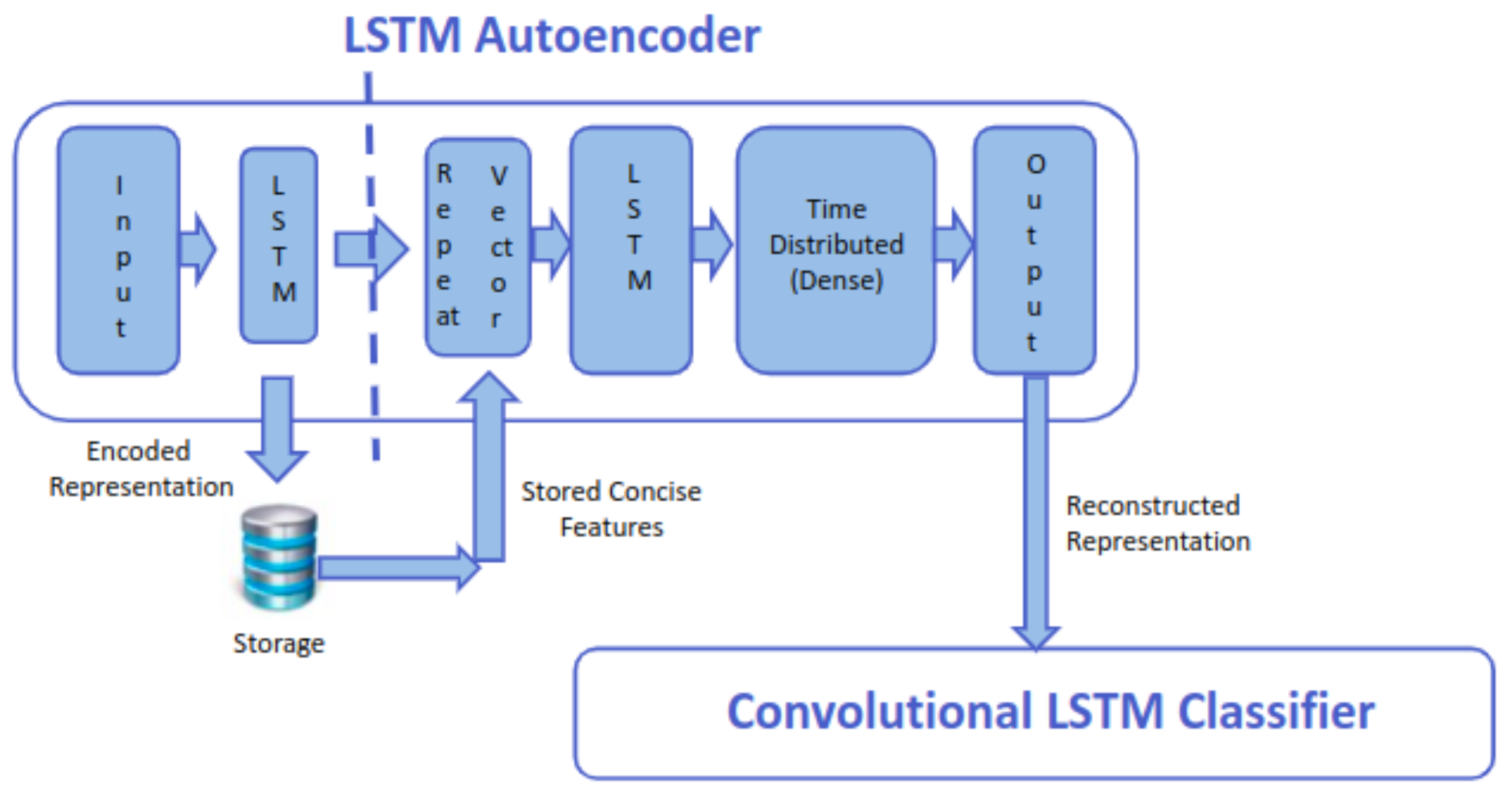}
\end{center}
\caption{LSTM autoencoder}
\label{Fig:LSTM_autoencoder}
\end{figure}

The activation function for all layers is ReLU, except for the last layer of the decoder, which uses linear activation function.  

The learning rate is set to 0.0001. Clip value is set to 0.5. For the Adam optimizer, if the clip value is not set, then a problem of exploding gradient appears, resulting in NaN values for the reconstruction loss. When the clip value is set to 0.5, that means the gradient value is also set to 0.5 for the cases where a gradient value is less than -0.5, or more than 0.5. MSE is used as the loss function. The reconstruction loss is 0.0221 after 300 epochs.

\item \textbf{Experiment 4:} 
The fourth experiment is performed with a convolutional LSTM deep autoencoder. The encoder consists of 5 layers in the following order: TimeDistributed (conv1D (64)) -> TimeDistributed (conv1D (64)) -> MaxPooling1D -> Flatten -> LSTM (100). 
\newline
The decoder consists of 8 layers in the following order: RepeatVector (4) -> LSTM (100) -> RepeatVector (4) -> Reshape -> TimeDistributed (conv1D (64)) -> TimeDistributed (conv1D (64)) -> TimeDistributed (conv1D (1)) -> TimeDistributed(Dense(9)). The activation function that is used by all of the conv1D and LSTM layers is ReLU, except for the last two layers in the decoder, which uses linear activation functions.

The default learning rate of ADAM optimizer that has been used is 0.001. Decay in the learning rate is set to 0.000001. MSE is used as the loss function, and the batch size is set to 16. The training/validation split ratio that has been used is 0.8/0.2. The validation loss is 0.0593 after 100 epochs.

\end{itemize}

\paragraph{Results}
The storage space is calculated using the st\_size attribute from the Python function os.stat(). The attribute returns the file size in Bytes. So it was divided by 1e+6 to determine the storage space in MB.

The results for the four experiments for the representation and the reconstruction components are discussed below. For all of the experiments, the storage size for the training data is 67.75MB.
\begin{itemize}
	\item \textbf{Experiment 1:}
	The saved representation storage size is reduced to 1.84MB.
	\item \textbf{Experiment 2:}
	The saved representation storage size is reduced to 60.228MB.
	\item \textbf{Experiment 3:}
	The saved representation storage size is reduced to 33.88MB.
	\item \textbf{Experiment 4:} 
	The saved representation storage size is reduced to 18.738MB.
\end{itemize}


Table \ref{tab1e:Reconstruction_loss} shows the results of the comparison of the storage reduction achieved for the different experiments. 

\begin{table}[t]
\caption{Storage Reduction for the UCI HAR data}
\label{tab1e:Reconstruction_loss}
\begin{center}
\begin{tabular}{p{50mm}p{20mm}p{30mm}p{20mm}}
 \toprule
\textbf{Dataset: UCI HAR data}  & \textbf{Reconstruction Loss (MSE)} &
\textbf{Accuracy (\%) on the classifier}  &\textbf{Storage Reduction (\%)}
\\ 
\hline \rule{0pt}{3ex}  
Exp. 1: MLP deep autoencoder & 0.0038 &24 & 90.18\\
Exp. 2: Convolutional deep autoencoder & 0.0024 & 95.28 & 11.18 \\
Exp. 3: LSTM autoencoder & 0.0221 &  52.01 & 49.99 \\
Exp. 4: Convolutional LSTM autoencoder & 0.0593 & 46.12 & 72.35 \rule{0pt}{3ex} \\
\bottomrule
\end{tabular}
\end{center}
\end{table}

\subsection{Discussion}
When the concise representation is taken from the encoder part of an autoencoder, it provides the option for reconstructing a pseudo-original data which is comparable to the original data. The classification performance of the reconstructed pseudo-original data depends on the size and dimension of the concise representation. For higher dimensions of the concise representation, the classification performance is higher as it has been demonstrated for convolutional autoencoder. The storage reduction for this specific case wasn't as good as the other options though. But if a good balance is found for the storage reduction, then this approach is better for meeting the requirements of both storage reduction and reconstruction of the pseudo-original data. 

\section{Threats to Validity} \label{Sec:Threats}
The approach has not been validated for real life streaming data yet. This work does not consider the issue of concept drift for the streaming data either. No ablation study has been done on the implemented models.

\section{Conclusion} \label{Sec:Conclusion}
Healthcare data is constantly growing, even more so in the recent years with the rise in remote health monitoring systems and the connected wearable sensor data. If efficient storage reduction techniques for the streaming data are not adopted, then response time to the remote health monitoring systems will eventually become slower. Our approach to storing the representative concise features rather than the whole incoming data provides a prospect to overcome the challenges raised by the growing need of the fast response time for the remote healthcare monitoring systems. Through our empirical study of the pattern reconstruction, we have found that the efficacy of the concise representation depended on the dimension and size of the representation. Further exploration and experimentation is needed to verify the preliminary findings found in this research work.



\medskip

\bibliography{neurips_cameraReady_2022}
\bibliographystyle{plainnat}

\end{document}